\documentclass{article}

\usepackage[preprint]{neurips_2023}

\usepackage[utf8]{inputenc} 
\usepackage[T1]{fontenc}    
\usepackage{url}            
\usepackage{booktabs}       
\usepackage{amsfonts}       
\usepackage{nicefrac}       
\usepackage{microtype}      
\usepackage{xcolor}         
\usepackage{makecell}  

\definecolor{linkColor}{rgb}{0.18,0.39,0.62}
\usepackage[colorlinks=true,linkcolor=linkColor,citecolor=linkColor,filecolor=linkColor,urlcolor=linkColor]{hyperref}       
\usepackage[most]{tcolorbox}
\usepackage{enumitem}
\usepackage{amsmath}
\usepackage{amssymb}
\usepackage{mathtools}
\usepackage{amsthm}
\usepackage{graphicx}
\usepackage{xspace}
\usepackage{pifont}
\usepackage{array}
\usepackage{bm}
\usepackage{ulem}
\usepackage{multirow}

\usepackage{rotating}
\usepackage{subcaption}
\usepackage{tablefootnote}
\usepackage{arydshln}
\usepackage{adjustbox}
\usepackage{wrapfig}
\usepackage{CJKutf8}
\usepackage{float}

\makeatletter

\newcommand{\Rmnum}[1]{\expandafter\@slowromancap\romannumeral #1@}
\makeatother

\title{VALL-E R: Robust and Efficient Zero-Shot Text-to-Speech Synthesis via Monotonic Alignment}

\author{%
 Bing Han$^1$\thanks{Work was done during internship at Microsoft Research Asia.} \ \ Long Zhou$^2$\thanks{Correspondence author} \ \ Shujie Liu$^2$ Sanyuan Chen$^2$ Lingwei Meng$^2$  \\ \textbf{Yanming Qian}$^1$ \textbf{Yanqing Liu}$^2$ \textbf{Sheng Zhao}$^2$ \textbf{Jinyu Li}$^2$  \textbf{Furu Wei}$^2$ \\
$^1$Shanghai Jiao Tong University\\
$^2$Microsoft Corporation
}

\begin{document}

\maketitle

\begin{abstract}
With the help of discrete neural audio codecs, large language models (LLM) have increasingly been recognized as a promising methodology for zero-shot Text-to-Speech (TTS) synthesis. 
However, sampling based decoding strategies bring astonishing diversity to generation, but also pose robustness issues such as typos, omissions and repetition. In addition, the high sampling rate of audio also brings huge computational overhead to the inference process of autoregression. 
To address these issues, we propose VALL-E R, a robust and efficient zero-shot TTS system, building upon the foundation of VALL-E \citep{wang2023neural}. Specifically, we introduce a phoneme monotonic alignment strategy to strengthen the connection between phonemes and acoustic sequence, ensuring a more precise alignment by constraining the acoustic tokens to match their associated phonemes.
Furthermore, we employ a codec-merging approach to downsample the discrete codes in shallow quantization layer, thereby accelerating the decoding speed while preserving the high quality of speech output. 
Benefiting from these strategies, VALL-E R obtains controllablity over phonemes and demonstrates its strong robustness by approaching the WER of ground truth. 
In addition, it requires fewer autoregressive steps, with over 60\% time reduction during inference.
This research has the potential to be applied to meaningful projects, including the creation of speech for those affected by aphasia.
Audio samples will be available at: \url{https://aka.ms/valler}.
\end{abstract}
%
%

\begin{figure*}[h!]
    \centering
    \includegraphics[width=11cm]{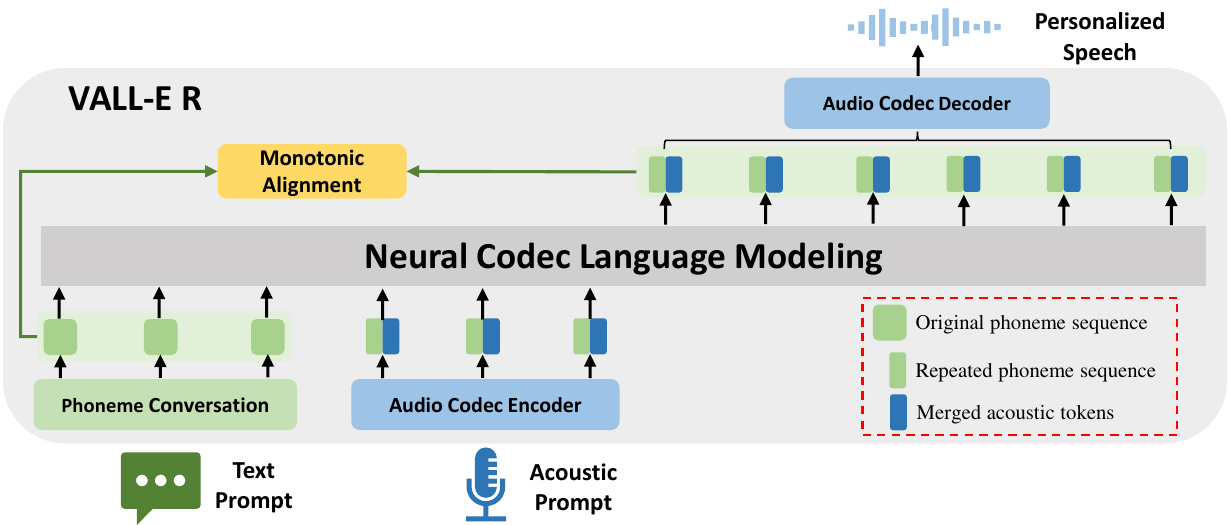}
    \caption{The overview of VALL-E R. It predicts both acoustic token (blue) and its corresponding phoneme sqeuence (green) synchronously, which can strengthen the alignment between phoneme and audio to improve the robustness of VALL-E model. 
    Note that VALL-E R achieves faster inference speeds by adopting the proposed merged codec codes within its autoregressive model.
    }
    \label{fig:overview}
\end{figure*}

\section{Introduction}
\label{sec:intro}
In recent years, Large Language Models (LLMs) have achieved impressive performance in various fields including natural language processing (NLP) tasks~\citep{achiam2023gpt,touvron2023llama} and computer vision (CV) tasks~\citep{ramesh2021zero, radford2021learning}. 
In audio domain, with the emergence of neural audio codecs~\citep{zeghidour2021soundstream, defossez2022high}, they can tokenize audio with high fidelity, making language modeling of audio modalities feasible~\citep{borsos2023audiolm}. 
For zero-shot text-to-speech (TTS) task, traditional methods often rely on speaker adaptation~\citep{chen2018sample,moss2020boffin} or speaker encoding~\citep{arik2018neural,cooper2020zero}, requiring additional fine-tuning, complex pre-designed features, or heavy structure engineering. And there is always a significant performance decline when facing unseen speakers. 
Leveraging the \textit{in-context learning} capabilities of LLM~\citep{brown2020language}, several works~\citep{wang2023neural, kharitonov2023speak, jiang2023mega, kim2023clam, hao2023boosting} attempt to apply LLM to model discrete audio tokens from neural codecs, and have shown amazing performance in zero-shot TTS, which can clone a timbre and prosody with just a few seconds of audio prompt. 

Although the TTS model based on LLM has achieved impressive results in maintaining naturalness and timbre, it still suffers from issues including robustness and efficiency when applied: (1) TTS task is a monotonic sequence to sequence task, but the decoder-only transformer structure in LLM only captures the monotonic association between phoneme sequences and audio through self-attention mechanism, which can easily lead to potential attention degradation problems when facing complex or long sequences, resulting in inaccurate generation results. (2) Neural codecs have a high sampling rate (75Hz in Encodec~\citep{defossez2022high}), and autoregressive can better model temporal information, bringing better results while also incurring significant computational costs in the inference stage. Recently, there have been some attempts to introduce phoneme information or transducer~\citep{graves2012sequence, gong2024advanced} to help solve robustness problems, but this has also brought about issues such as high inference consumption~\citep{song2024ella} and slow training speed~\citep{du2024vall}. Some other works attempt to improve efficiency by changing the pattern of token organization, but the effect is not very obvious~\citep{borsos2023audiolm, wang2023neural, copet2024simple,yu2024megabyte}. 

In this paper, we introduce VALL-E R, a robust and efficient neural codec language model for zero-shot TTS, aiming to alleviate the issues of robustness and inference efficiency encountered by its predecessor VALL-E \citep{wang2023neural}. 
As shown in Figure~\ref{fig:overview}, VALL-E R incorporates phoneme prediction into acoustic token prediction during the training process and utilize monotonic alignment strategy to guide the inference, with almost no computational burden introduced throughout the entire process. 
Moreover, VALL-E R employs a novel codec-merging method that enables the use of condensed codec codes within its autoregressive model, greatly improving the inference speed via reducing the number of autoregressive steps in the inference stage.
In summary, our proposed VALL-E R provides the following key contributions:
\begin{itemize}
    \item VALL-E R employs a codec-merging approach, which can reduce the sample rate of codes in first layer without affecting the generated speech quality and retraining the codec model, thereby significantly improving the inference speed of neural codec LM-based TTS model.
    \item VALL-E R introduces phoneme monotonic alignment strategy, which can enhance the alignment between phonemes and acoustic sequence to improve the robustness of decoder-only Transformer TTS.
    \item VALL-E R exhibits stronger controllability in prosody due to its phoneme-based input mechanism in inference. Experimental results demonstrate that VALL-E R can clone timbre while adjusting pronunciation prosody, achieving the goal of voice conversion.
\end{itemize}

VALL-E R is a purely reseach project with no immediate plans to transition it into a commercial product or to broaden public access. This technology has the potential to create synthesized voices that reflect individual identities, lending itself to a variety of applications such as education, entertainment, journalism, personal content creation, assistive technologies, automated voice systems, language translation, and chatbots. The ability of VALL-E R to replicate a specific person's voice convincingly varies based on the length and clarity of the provided speech sample, any present background noise, and other influencing elements. There are inherent risks associated with the technology, including the possibility of misusing it to mimic voice recognition systems or impersonate individuals. Our experiments were conducted with the presumption of consent from the individuals whose voices were being reproduced. In any future applications involving speakers not initially included in the model, measures must be taken to ensure consent for the use of their voice, along with the implementation of tools for detecting synthetic speech. Should there be any concerns about VALL-E R being utilized inappropriately, illegally, or in a manner that infringes upon personal or others' rights, reports can be submitted through the designated Report Abuse Portal.

\section{Related Work}
\subsection{Neural Codec Language Modeling}
Recently, language models have increasingly garnered attention within both the academic and industrial sectors, demonstrating their formidable capabilities in feature modeling and problem-solving across domains such as text~\citep{touvron2023llama}, images~\citep{ding2021cogview}, and videos~\citep{kondratyuk2023videopoet}. In the audio domain, 
AudioLM~\citep{borsos2023audiolm} has been trained on discrete audio tokens, accomplishing audio synthesis tasks through hierarchical prediction of these tokens, and text conditional control can also be achieved through models such as CLAP~\citep{zhao2023clap}. 
For zero-shot TTS, VALL-E~\citep{wang2023neural} pioneered the integration of discrete codecs and approached TTS as a conditional language modeling task, enabling voice generation in unseen speakers from a mere short speech sample, thereby underscoring the powerful representational capacity of language models in TTS. 
Following it, numerous improvements have been made. VALL-E X~\citep{zhang2023speak} expanded upon multilingual capabilities, supporting a broader range of tasks. VioLA \citep{wang2023viola} further integrated various cross-modal tasks involving speech and text into a single codec-based Transformer decoder-only model.
In terms of robustness, EALL-V~\citep{song2024ella} enhanced robustness by implicitly establishing a connection between text and audio during the synthesis process through text insertion. VALL-T~\citep{du2024vall} improved robustness by combining with a Transducer and introducing text control by adjusting relative position encoding. Meanwhile, CLaM-TTS~\citep{kim2023clam} leveraged a pre-trained language model and probabilistic discrete learning to enhance the expressiveness of synthesized speech, marking significant advancements in the field. 
Concurrently with our work, RALL-E \citep{xin2024rall} attempts to enhance  the robustness of VALL-E through the adoption of chain-of-thought (CoT) prompting techniques.
Although the above works have effectively improved the performance of decoder-only transformer based text-to-speech system, no one has yet attempted to improve the efficiency of autoregression, which is currently a pain point.

\subsection{Monotonic Alignment}
In autoregressive acoustic modeling, issues such as word skipping, repetition, and attention collapse often occur due to inaccuracies in the attention alignments learned within the encoder-decoder framework.
To alleviate this problem, considering some properties of the alignments between text and waveform sequence, applying monotonic mechanism to enhance attention has demonstrated to be particularly effective to strictly preserve monotonicity and locality~\citep{he2019robust}. 
Considering the alignments between text and speech are depending on their positions, numerous TTS models have incorporated location-based attention mechanisms to exploit the positional information for more accurate alignment~\citep{sotelo2017char2wav, vasquez2019melnet, battenberg2020location}. 
For monotonic attention, it leverages the prior that the alignments between text and speech are monotonic~\citep{he2019robust, raffel2017online}. 
In each decoding step, the attention alignment position moves forward at most one step, so there is no situation where any input unit is skipped or repeated. 
While such methods have enhanced stability by introducing linear alignment, the constraints of hard attention significantly hinders the encapsulation of global contextual information, where the model only focuses on one token at a time.
Therefore, models in~\citep{tachibana2018efficiently, chiu2017monotonic} restrict the attention on the source sequence into a sliding window, replacing hard attention.
Furthermore, Chen et al.~\citep{chen2020multispeech} introduce an innovative approach to penalize off-diagonal attention weights by constructing a band mask that promotes the concentration of attention weights within a diagonal band, thus enhancing alignment fidelity while maintaining contextual awareness.
Although these methods have effectively improved the robustness of autoregressive TTS systems, they are all based on the encoder-decoder architecture and are not compatible with the current popular decoder-only architecture.

\section{VALL-E R}
In this study, we propose a robust and efficient zero-shot TTS system named VALL-E R. We first introduce the codec-merging approach in Section \ref{sec:merge_codec} which can improve inference speed without retraining the codec, and then illustrate the monotonic alignment strategy in decoder-only neural codec LM in Section \ref{sec:monotonic_alignment}.

\begin{figure}[h!]
		\centering
		\includegraphics[width=12cm]{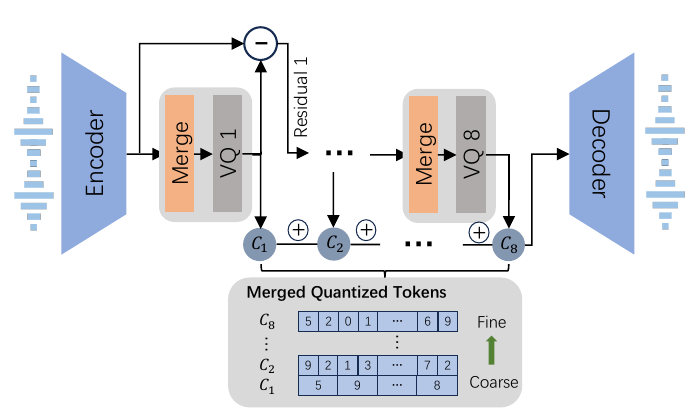}
	\caption{The architecture of proposed codec-merging approach. VQ denotes vector quantizer layer here. Through the codec-merging module, the code in the first layer is downsampled twice, meaning that every two adjacent codes in the first layer are the same and the other layers remain unchanged. }
	\label{fig:codec}
\end{figure}

\subsection{Codec-Merging Approach}
\label{sec:merge_codec}
Building upon the foundational work of Encodec~\citep{defossez2022high}, we introduce the concept of a merged codec. This innovative approach enables the downsampling of discrete codes across various layers through alterations in the inference forward process. Remarkably, this is achieved without necessitating any retraining or fine-tuning of the model, presenting a significant advancement in the efficient manipulation of audio data representations. 

The visual description of the proposed codec can be seen in Fig.~\ref{fig:codec}. 
The overall architecture of the model remains consistent with Encodec, which comprises of three components: 1) a convolution-based encoder that maps waveform data $x^{1\times L}$ into a sequence of latent representations $z^{F\times T}$, where $F$ is channel and $T$ is length of extracted codes; 
2) a decoder that reconstructs the data $\hat{x}^{1\times L}$ from a sequence of the quantized latent representations $\hat{z}^{F \times T}$; 
3) an 8-layer residual vector quantizer (RVQ) module, which can convert the continual latent vector $z^{F\times T}$ into the discrete code representation $C^{8\times T}$ iteratively. The main difference is that our merged codec inserts a codec-merging module before the vector quantizer module to downsample the representation $z^{F\times T}$.

Assuming the merged rate of layer $d$ is $m_d$, $r_d^{F\times T}$ represents the residual input of layer $d$. Then the codec-merging module consists of two steps: the first one is downsampling the residual input $r_d^{F\times T}$ to $r_{md}^{F\times (T/m_d)}$ by average pooling, and then upsampling $r_{md}$ to its original length through repeat operation. 
Next, the residual representation processed by the Merge module will be feed into the following VQ layer to quantized into discrete code $C_d^{1 \times T}$ through nearest-neighbour lookup over the codebook embeddings. Through the merge module, we reduced the resolution of $C_d^{1 \times T}$ by ensuring consistent code of consecutive $m_d$ frames.

\subsection{Neural Codec LM with Monotonic Alignment}
\label{sec:monotonic_alignment}
Previously, monotonic strategies were only applicable to encoder-decoder structures. To address the robustness issue in the decoder-only transformer based TTS, we integrated phonemes prediction into the training of neural codec LM and design the monotonic alignment stratege during the inference process. The overview is illustrated in Fig.~\ref{fig:model} and details of training and inference are discussed in the subsequent sections.

\subsubsection{Training with Phoneme Prediction}
To achieve a good trade-off between speech quality and inference speed, our VALL-E R includes two models: autoregressive (AR) and non-autoregressive (NAR), which is following VALL-E~\citep{wang2023neural}. Specifically, given a training data pair $\{\mathbf{s}, \mathbf{p}\}$, where $\mathbf{s}$ is speech sample, and $\mathbf{p}=\{p_1, p_2, \dots, p_L\}$ is its corresponding phoneme transcription. Then, the codec-merging model introduced in Sec.~\ref{sec:merge_codec} is utilized to compress speech waveform $\mathbf{s}$ into discrete acoustic tokens $\mathbf{A}$ with 8 quantizers, formulated as: $\mathrm{MergeCodec}(\mathbf{x})=\mathbf{A}^{8\times T}=\{\mathbf{a}^1, \dots, \mathbf{a}^8\}$, where $T$ is length of discrete codes and $\mathbf{a}^i=\{a_1, \dots, a_T\}$ represent the tokens in the $i$-th layer. Because the training of VALL-E R requires the aligned phonemes and acoustic tokens, aligner tool is adopted here to align $\mathbf{p}$ with acoustic tokens $\mathbf{A}$, denoted as $\hat{\mathbf{p}}_{1:T}=\{\hat{p_1}, \hat{p_2}, \dots, \hat{p_L}\}$ where $\hat{p_i}$ contains $N_i$ repetitions of  $p_i$ and $\sum_{i=1}^{L}N_i=T$. 

\begin{figure*}[t]
		\centering
		\includegraphics[width=13cm]{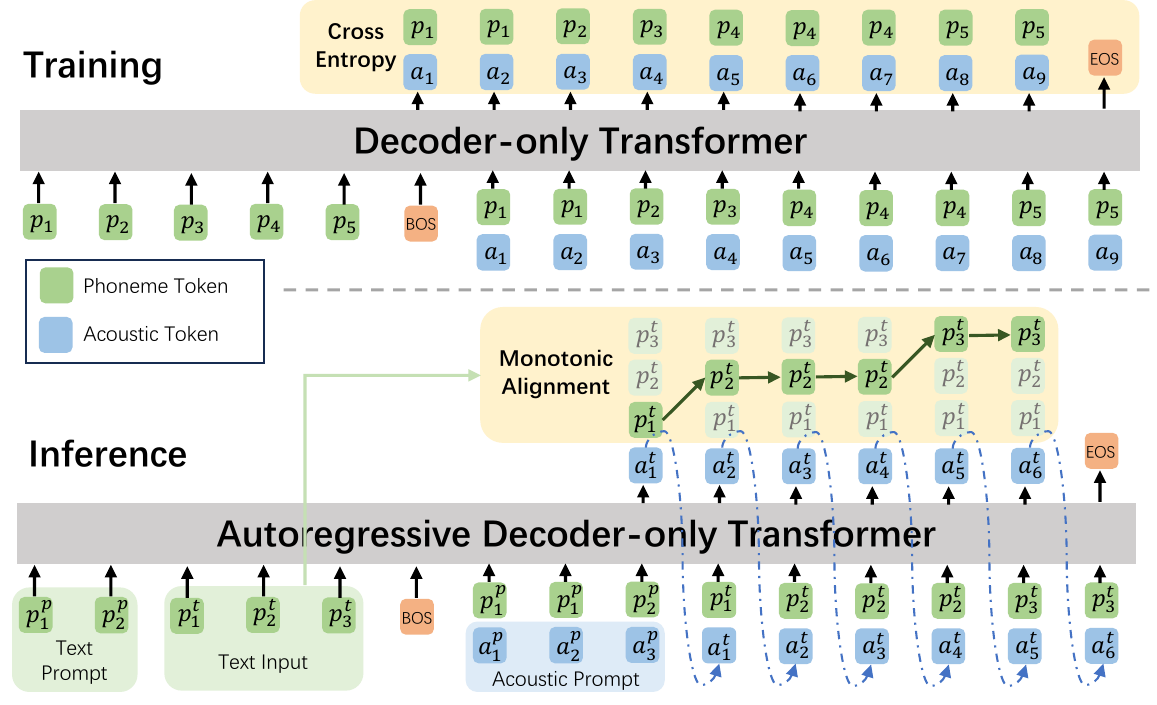}
	\caption{The training and autoregressive inference process of proposed VALL-E R. In the training process, it enhances alignment between phonemes and acoustic by incorporating phoneme prediction with teacher-forcing. And the inference process will generate the sequences monotonically based on phonemes of text input.}
	\label{fig:model}
\end{figure*}

For AR stage, to enhance the connection between phoneme and acoustic sequence, we build a neural codec LM $\theta_{AR}$ to model the discrete acoustic tokens $\mathbf{a}^1_{1:T}$ from the first quantizer of codec-merging model with phoneme prediction. As shown in Fig.\ref{fig:model}, it's conditioned on the phoneme sequence $\mathbf{p}$ to generate both the acoustic token $\mathbf{a}^1_{1:T}$ and aligned phonemes $\hat{p}_{1:T}$ simultaneously, formulated as maximizing the following probability:
\begin{align}
    & p(\mathbf{a}^1_{1:T}, \hat{\mathbf{p}}_{1:T} | \mathbf{p}; \theta_{AR}) \\
    = & \prod_{t=1}^{T} p(a_t, p_t|a_{1:t-1},\hat{\mathbf{p}}_{1:t-1},\mathbf{p}_{1:L};\theta_{AR})
\end{align}

In the second stage, we train a NAR LM $\theta_{NAR} $to generate the acoustic tokens from 2$nd$ to 8-$th$ layer quantizers iteratively. It's conditioned on phoneme sequences $\mathbf{p}$, the previous few layers of generated acoustic tokens $\mathbf{a}^{1:n}$ and phonemes alignment $l_{1:T}$ to predict next layer of acoustic token $\mathbf{a}^{n+1}$, formulated as maximizing:
\begin{align}
    & p(\mathbf{a}^{2:8}_{1:T} | \hat{\mathbf{p}}_{1:T}, \mathbf{p}_{1:L}; \theta_{NAR}) \\
    = & \prod_{n=2}^{8} p(\mathbf{a}^{n}_{1:T}|\hat{\mathbf{p}}_{1:T}, \mathbf{a}^{1:n-1}_{1:T}, \mathbf{p}_{1:L};\theta_{NAR})
\end{align}
We also share the parameters of the acoustic embedding layer and the output prediction layer, which means the weights of the $j$-th prediction layer are the same as the $(j + 1)$-th acoustic embedding layer.

\subsubsection{Inference with Monotonic Alignment}
\label{sec:inference}
After training with teacher forcing, the neural codec LM we obtained is surprising in context learning ability. With just a 3 seconds acoustic prompt, we can replicate the timbre and prosody of unseen speakers without any fine-tuning or adaptation methods~\citep{wang2023neural}. 
Take Fig.~\ref{fig:model} as an example of autoregressive inference, we convert the text input into phonemes sequence $\mathbf{p}^t=\{p_1^t, \dots, p_3^t\}$ and concatenate it with transcription phoneme of acoustic prompt $\mathbf{p}^p=\{p_1^p, p_2^p\}$ to form phoneme tokens for model. In the following, we will use acoustic tokens $\mathbf{a}=\{a_1^p, \dots, a_n^t\}$ and corresponding aligned phoneme tokens $\mathbf{p}^a=\{p_1^p, \dots, p_n^t\}$ as condition to predict the audio and phoneme of next step autoregressively.

In order to improve the robustness of the model, we adopted the Monotonic Alignment (MA) strategy during the inference process, which means that the predicted phone needs to be aligned with the text input $\mathbf{p}^t$ and can only maintain the current or jump to the next phoneme. Specifically, at each step $i$, the current input phoneme token is $p_j^t$. The output representation is $\mathbf{e_i}$, and corresponding probability of phoneme $p_j^t$ and $p_{j+1}^t$ is denoted as $e_{i,j}$ and $e_{i,j+1}=1-e_{i,j}$ respectively. Then, the phoneme pointer would decide to keep $p_j^t$ unmoved or jump to $p_{j+1}^t$ by sampling:
\begin{equation}
    z_{i,j} \sim Bernoulli(\frac{1}{1+exp(e_{i,j})}) 
\end{equation}
The sampled phoneme will be used as input for the $i+1$ step and this AR process will repeat until all phonemes have been covered. To increase the diversity of synthesis process, sampling based decoding is used for acoustic tokens prediction. And for NAR model, we adopt greedy search method to generate the $(j+1)$-th layer acoustic tokens based on the self-aligned phoneme sequence and previous few layers of generated acoustic tokens. Finally, decoder of neural codec is adopted here to convert generated discrete codes into waveform. In summary, our inference process has the following three characteristics~\citep{tan2021survey} by using MA strategy:
\begin{itemize}
    \item \textit{Locality}: Each phoneme token can correspond to one or several consecutive acoustic tokens, ensuring a flexible and accurate mapping. Conversely, each acoustic token is uniquely aligned to a single phoneme token. This one-to-one alignment strategy effectively prevents issues such as misreading, enhancing the clarity and stability of the model. 
    \item \textit{Monotonicity}: If phoneme $p_a$ follows $p_b$ in the phonemes sequence, the corresponding acoustic tokens for $p_a$ are also positioned sequentially after those for $p_b$. This sequential alignment is critical as it inherently prevents the repetition of words.
    \item \textit{Completeness}: It is mandated that each phoneme token is represented by at least one corresponding acoustic token. This coverage requirement is essential for preventing the word skipping.
\end{itemize}

\subsubsection{Inference for Controlling Prosody}
In the inference process, benefiting from the powerful in context learning ability of the LM, our proposed VALL-E R can automatically clone both timbre and prosody of the speaker in the prompt by predicting the acoustic and phoneme autoregressively. Because VALL-E R explicitly models phoneme, it has strong control over prosody: when we use preset phoneme sequences to replace the self-predicted phoneme sequences in the inference process, we can use the preset prosody to generate speech, thereby achieving the effect of controlling prosody and timbre separately. 
It can also be regarded as a voice conversion task, whose goal is to make the timbre of target speech sound like that of prompt speech without changing the linguistic information and prosody of source speech.

\section{Experiments Setup}

\subsection{Dataset}
We use LibriSpeech~\citep{panayotov2015librispeech} dataset in our experiments, which contains 960 hours of multi-speaker transcribed English speech data. Because VALL-R requires phoneme alignment information during training and inference processing, Montreal Forced Aligner (MFA)~\citep{mcauliffe2017montreal} is utilized to align transcript and audio. For the neural codecs, we use Encodec~\citep{defossez2022high}, which applies Residual Vector Quantization (RVQ) to compress 24khz waveform into 8-layers discrete codes at 75hz. To synthesize higher quality audio, we adopt Vocos~\citep{siuzdak2023vocos} as vocoder which is aligned with Encodec.

\subsection{Evaluation Metrics}
\textbf{Objective Metrics} We evaluated robustness, speaker similarity, sound quality and efficiency with several objective metrics. For robustness and intelligibility, we measure them by computing the word error rate (WER) of the synthesized transcription from generated speech concerning the input text, with Conformer-Transducer ASR model\footnote{https://huggingface.co/nvidia/stt\_en\_conformer\_transducer\_xlarge}~\citep{gulati2020conformer}. 

For speaker similarity, we employ WavLM-TDNN model\footnote{https://huggingface.co/microsoft/wavlm-base-plus-sv} \citep{chen2022wavlm, chen2022large} which extract the speaker vector representing the voice attribute of the speakers. We measure the speaker similarity by computing the cosine similarity between the generated and the prompt speeches, which is denoted as Spk-Sim. 

To evaluate the sound quality of the reconstruction audio, we adopt the metrics from speech enhancement fields, such as the Perceptual Evaluation of Speech Quality (PESQ) and Short-Time Objective Intelligibility (STOI) to evaluate the performance.

\textbf{Subjective Metrics} We measure the quality of the generated speech from human evaluations via three types of Mean Opinion Score (MOS): 1) Quality MOS (QMOS) for the speech quality assessment, 2) Similarity MOS (SMOS) to measure speaker similarity between the prompt and the generated speech, 3) Comparative MOS (CMOS) to assess the overall naturalness and quality of the synthesized speech against the baseline, which is ranging from -1 to 1.

\subsection{Baseline}
To demonstrate the robustness and inference efficiency of our system, we selected multiple classic baseline systems for comparison. For robustness, besides its predecessor VALL-E~\citep{wang2023neural}, we mainly compare two recent works which are proposed to improve robustness. One is VALL-T, it adopts transducer to enhance text controllability via adjusting relative position encoding. And another one is ELLA-V~\citep{song2024ella}, which interleaves sequences of acoustic and phoneme tokens to achieve the fine-grained control at the phoneme level. To ensure fair comparison, all the systems including VALL-E R and baselines, are trained with the same dataset (LibriSpeech) and architecture (stacked 12-layers Transformer). As for efficiency, since the inference time of neural codec LM based TTS system mainly depends on the number of autoregressive steps, we selected representative methods including AudioLM~\citep{borsos2023audiolm}, VALL-E~\citep{wang2023neural}, ELLA-V~\citep{song2024ella}, RALL-E~\citep{xin2024rall}, MusicGen~\citep{copet2024simple}, and compared the impact of different token arrangement methods on efficiency.

\subsection{Training Configuration}
During the training, we trained the autoregressive (AR) model in the teacher-forcing manner and selected the first 3 seconds as the prompt to optimize the non-autoregressive (NAR) model. 
Regarding the model architecture, both our AR and NAR models are structured with a 12-layer Transformer framework. Key parameters including the number of attention heads, embedding dimension, hidden state dimension, feed-forward network dimension, and dropout rate are configured to 16, 1024, 1024, 4096, and $0.1$, respectively. These models are trained on 8 NVIDIA V100 16GB GPUs across 400k steps utilizing the AdamW optimizer. The training protocol initiates with a learning rate warm-up phase over the first 32k updates, reaching a peak at $5 \times 10^{-4}$. Subsequently, the learning rate undergoes a linear decay coupled with a weight decay factor of $0.01$.


\section{Results and analysis}
\subsection{Zero-shot Text-to-Speech tasks}
We measure the performances of the proposed VALL-E R under two different tasks following the configuration used in~\citep{wang2023neural, kim2023clam} on LibriSpeech test-clean ranging from 4 to 10 seconds: 
1) \textit{Continuation}: we use the first 3 seconds of the utterance and the corresponding text transcription as prompt respectively, and ask model to synthesize the subsequent portion of the speech seamlessly.
2) \textit{Cross-sentence}: Given the target text, a 3-second prompted speech, and its corresponding transcript, which contains a different sentence from the target text, the task is to synthesize the speech with target text following the style of the provided speech prompt.
\begin{table}[h]
  \caption{Objective performance comparison on the \textit{continuation} and \textit{cross-sentence} zero-shot TTS. $^{*}$ denote the results quoted from the original paper, and others are reproduced. Encodec represents that ground-truth audios were passed through the encoder and decoder of Encodec.}
  \label{tab:results_overview}
  \centering
  \begin{tabular}{lcccc}
    \toprule
    \multirow{2}{*}{Model} &\multicolumn{2}{c}{Continuation} & \multicolumn{2}{c}{Cross-sentence} \\
         & WER $\downarrow$  & Spk-Sim $\uparrow$ &  WER $\downarrow$  & Spk-Sim $\uparrow$ \\
    \midrule
    \midrule
    Ground Truth & 1.41  &   0.923 & - & - \\
    Encodec & 1.62  &   0.913  & - & - \\
    \midrule
    VALL-E~\citep{wang2023neural}  & 2.37 & 0.875  & 5.48 & \textbf{0.975}  \\
    VALL-T~\citep{du2024vall}*  &  - &   -  &  4.16 &  - \\
    ELLA-V~\citep{song2024ella}*   &  2.28 & 0.870 & - & - \\
    ELLA-V~\citep{song2024ella}   &  2.10 & 0.856 & 7.15 & \textbf{0.975}\\
    \midrule
    VALL-E R & \textbf{1.58} &  \textbf{0.876} & \textbf{3.18} & 0.974\\
    \bottomrule
  \end{tabular}
\end{table}

\paragraph{Objective Evaluation} We list the objective evaluation results of \textit{continuation} and \textit{cross-sentence} in Table~\ref{tab:results_overview}. 
Under the continuation configuration, although ELLA-V achieves a certain improvement over the VALL-E via connecting phoneme tokens with their corresponding acoustic tokens, our proposed VALL-E R employs a monotonic alignment strategy for more precise control and approaches the performance of Encodec in terms of Word Error Rate (WER), which can be considered as upper bound. 
In the cross-sentence task, due to the discontinuity of content, which mismatches the continuation training manner, all systems experience varying degrees of degradation in WER. Among these, ELLA-V, suffers a more significant impact because of its strategies such as local advance. But VALL-E R is still the best in WER, reflecting the robustness and intelligibility of our model. In terms of speaker similarity (Spk-Sim), VALL-E R is comparable to VALL-E and ELLA-V, indicating that although our method explicitly encodes text contents, it does not affect the performance of timbre cloning. 

\begin{table}[h]
  \caption{Subjective performance for MOS score on LibriSpeech test clean. QMOS and SMOS scores include a 95\% confidence interval. CMOS denotes comparision with baseline system VALL-E.}
  \label{tab:res_mos}
  \centering
  \begin{tabular}{lccc}
    \toprule
    Model     & QMOS & SMOS & CMOS\\
    \midrule \midrule
    Ground Truth & $4.22\pm 0.11$  & $4.18\pm 0.15$ & +0.33\\
    VALL-E & $3.96 \pm 0.18$  & $3.84 \pm 0.21$ & 0.00\\
    VALLE-R & $4.02 \pm 0.20$ & $3.89 \pm 0.16$ & +0.07\\
    \bottomrule
  \end{tabular}
\end{table}

\paragraph{Subjective Evaluation} Table~\ref{tab:res_mos} showcases the results of subjective audio evaluations. 
VALL-E R outperforms the baseline (VALL-E), in quality and intelligibility, as indicated by QMOS. 
According to SMOS measurements, our compliance with prompt audio is comparable to baseline.
The comparative scores (CMOS) highlight VALL-E R’s proximity to the Ground Truth regarding naturalness, clarity, and comprehensibility. 
Overall, VALL-E R’s generated speech exceed the baseline in naturalness, quality and intelligibility, which is consistent with the results of objective evaluation.


\subsection{Controllability of VALL-E R}

\begin{table}[h]
  \caption{Performance for the \textit{cross-sentence} prosody controlling tasks. The lower the score, the higher the similarity of the prosody.}
  \label{tab:mcd}
  \centering
  \begin{tabular}{lccc}
    \toprule
    Model     & MCD-DTW-SL $\downarrow$ \\
    \midrule \midrule
    VALL-E~\citep{wang2023neural}  & 9.03  \\
    ELLA-V~\citep{song2024ella} & 11.77 \\
    VALL-E R & 8.55 \\ \midrule
    VALL-E R-Prosody &  \textbf{7.82} \\
    \bottomrule
  \end{tabular}
\end{table}

Benefiting from the explicit modeling of phonetic sequences, VALL-E R is capable of precisely controlling the duration of each phoneme's articulation, thereby achieving timbre conversion while preserving prosody. In this experiment, we randomly selected an audio sample from the LibriSpeech test set to serve as acoustic prompt, providing the reference timbre. Another audio sample was randomly chosen, with its phonetic transcription serving as both the textual prompt and prosody reference. We conducted a comparison among multiple baseline systems, calculating the Mel Cepstral Distortion with Dynamic Time Warping and weighted Speech Length (MCD-DTW-SL)~\citep{chen2022v2c} between the generated audio and the target audio to evaluate the similarity in speech length and alignment quality as a measure of prosody. Results are presented in Table~\ref{tab:mcd}, neither VALL-E nor ELLA-V is capable of precise duration control, thus limiting their speech generation capabilities to the prosody of the given acoustic prompt. For VALL-E R-Prosody, we can employ the aligned phoneme sequence from the target prosody audio as input for the neural codec language model. This enables the preservation of target prosody while cloning the timbre from the acoustic prompt. According to the results, this approach also achieved the lowest MCD-DTW-SL, indicating a superior preservation of the prosody. It also indicates that our VALL-E R has better controllability.

\subsection{Efficiency Comparision}

\begin{table*}[h]
  \caption{Performance of inference speed. The inference time indicates the generation time of 10s speech. These results are all measured based on 75Hz Encodec~\citep{defossez2022high}. Avg. denotes Average. AR and NAR represent autoregressive and non-autoregressive model respectively. Results with $^{*}$ are quoted from CLaM-TTS~\citep{kim2023clam}. According to statistics in the test set, the average 10 second speech corresponds to 105 phonemes.}
  \label{tab:efficiency}
  \centering
  \begin{adjustbox}{width=1.0\textwidth,center}
  \begin{tabular}{lcccc}
    \toprule
    Model     & Avg. AR Steps & Avg. AR Time & Avg. NAR Time & Avg. Inference Time \\
    \midrule \midrule
    AudioLM~\citep{borsos2023audiolm} & 750 * 8  &  $>$ 40 & -  &  $>$ 40 \\ 
    VALL-E~\citep{wang2023neural}   & 750  &  10.1246 & \textbf{0.1478} &  10.2724    \\
    ELLA-V~\citep{song2024ella}   & $\sim$ 105 * 2 + 750 &  15.5581 &  0.1993  & 15.7574   \\
    RALL-E~\citep{xin2024rall} & $\sim$ 105 + 750 & 12.1152 & 0.1683 & 12.2835 \\
    MusicGen~\citep{copet2024simple} & 7 + 750 & 10.2693 & - & 10.2693\\
    \midrule
    VALL-E~\citep{wang2023neural} & - & - & - & 6.2$^{*}$ \\
    VoiceBox~\citep{le2024voicebox} & - & - & - & 6.4$^{*}$(64 NFE) \\
    CLaM-TTS~\citep{kim2023clam} &  - & - & - & 4.15$^{*}$ \\
    \midrule
    VALL-E R (2x)  & \textbf{375} & \textbf{3.5251} & 0.1489  &  \textbf{3.6740} \\
    \bottomrule
  \end{tabular}
  \end{adjustbox}
\end{table*}

We conducted a comparison of the inference times required to generate 10 seconds of speech among several neural codec language model inference patterns, detailing the time needed for AR and NAR module in Table~\ref{tab:efficiency}. 
The first row is the flatten pattern in AudioLM, which generates all 8 layers of discrete audio codecs autoregressively, necessitating 750*8 steps to generate 10 seconds of audio, resulting in extremely slow speeds. 
VALL-E adopts a coarse-to-fine pattern, generating the first layer autoregressively, then employing non-autoregressive module for the codec of subsequent 7 layers, significantly reducing inference time. 
ELLA-V, building upon VALL-E, introduces a large number of phoneme tokens (around 2*105 in 10s speech) into the autoregressive process. While this enhances robustness, it considerably slows down inference efficiency. Similarly, RALL-E introduces control information for duration and pitch in autoregression through \textit{Chain-of-Thought} prompting, which also increases inference time. 
MusicGen, on the other hand, is based on a delay pattern. Despite discarding the non-autoregressive module, its speed does not show a significant improvement over VALL-E. 

By employing our proposed codec-merging method, which reduces the bitrates at the first layer of RVQ from 75hz to 37.5hz, we can effectively decrease the number of autoregressive steps required, thereby significantly enhancing inference speed. 
Additionally, due to the exponential complexity of the Transformer's self-attention mechanism, halving the sampling rate results in a speed increase of more than two times. This method also outpaces both the flow-matching based NAR model VoiceBox and the Mel-VAE based CLaM-TTS in terms of speed. 

\subsection{Analysis and Discussion}

\begin{table*}[h]
  \caption{Effectiveness of codec-merging method. Experiments are conducted on LibriSpeech test set.}
  \label{tab:results_codec}
  \centering
  \begin{tabular}{lcccccc}
    \toprule
    Model   & Merge &  No Merge & Merge Rate & PESQ(NB) & PESQ(WB) & STOI \\
    \midrule
    Encodec   &  - & 1-8 &  - & 3.6188 & 3.2427 & 0.9498 \\
       &  1 & 2-8 &  2 & 3.5686 & 3.1742 & 0.9468 \\
       &  1-4 & 5-8 & 2 & 3.1820 & 2.7046 & 0.9274 \\
       &  1-8 & - &  2 & 2.0160 & 1.4901 & 0.8323 \\
       &  1 & 2-8 &  3 & 3.5294 & 3.1206 & 0.9449 \\
       &  1 & 2-8 &  4 & 3.4945 & 3.0805 & 0.9436 \\
    \bottomrule
  \end{tabular}
\end{table*}

\paragraph{Effect of Merged Codec}
The codec-merging strategy results in a reduction of the model's bitrate, thus we have explored the impact of different configurations of the merged codec on audio quality, presenting the results in Table~\ref{tab:results_codec}. These experiments involved reconstruction tasks performed on the LibriSpeech test set, where the quality of audio was assessed by comparing the ground truth with the reconstructed audio using metrics PESQ and STOI. 
Results reveal that when we only reduce the sampling rate by half for the first layer codec, the audio quality and intelligibility are almost unaffected. However, as we expand this reduction to 4 layers and 8 layers, a significant decrease in performance is observed, which does not enhance the inference speed. Thus, downsampling only the first layer achieve the trade-off between quality and efficiency. Although all our experiments are conducted with 2x downsampling rate, we also explored the effects of increasing the downsampling to 3x and 4x. It was found that tripling and quadrupling the downsampling rate only resulted in a slight decrease in audio quality, which underscores the potential of VALL-E R to further improve inference efficiency.

\begin{figure}[h]
		\centering
		\includegraphics[width=12cm]{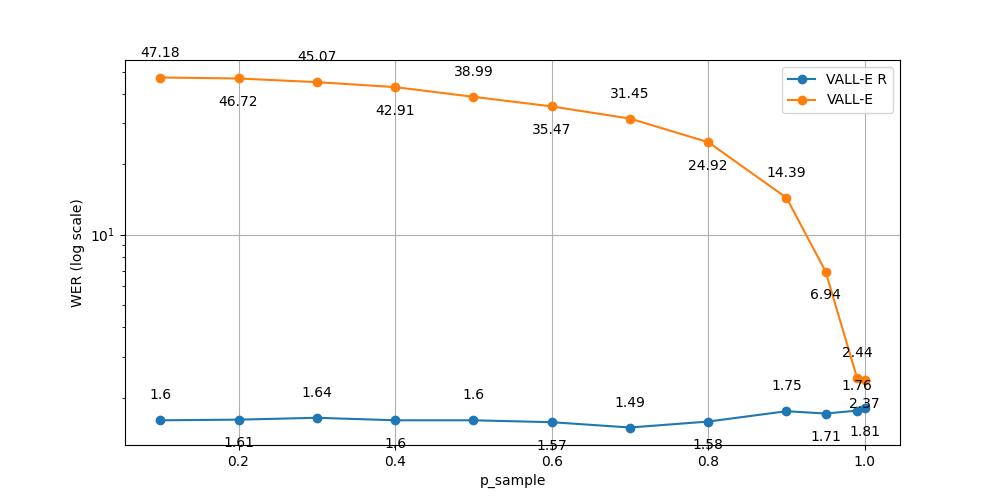}
	\caption{The trends of WER, with respect to the variations in top\_p value when decoding.}
	\label{fig:res}
\end{figure}

\paragraph{Effect of top\_p in Inference}
In order to demonstrate the stability and robustness of our proposed VALL-E R, we conducted ablation experiments during the decoding stage, attempting to analyze the impact of the top\_p in sampling on WER. And the figure of WER changing with top\_p is shown in Fig.~\ref{fig:res}. For VALL-E, due to its lack of control over text, reducing the top\_p parameter leads to a loss of randomness in the model. This makes the model inclined to predict silent frames to maximize the overall probability of the generated content, resulting in an endless loop and consequently poor Word Error Rate (WER). However, by incorporating a monotonic alignment strategy, our proposed VALL-E R becomes possible to finely control the pronunciation text. Therefore, even when reducing top\_p, VALL-E R not only maintains diversity but also ensures the normal generation of content, exhibiting strong robustness.

\begin{figure*}[htb]

	\begin{minipage}[b]{1.0\linewidth}
		\centering
		\centerline{\includegraphics[width=14cm]
  {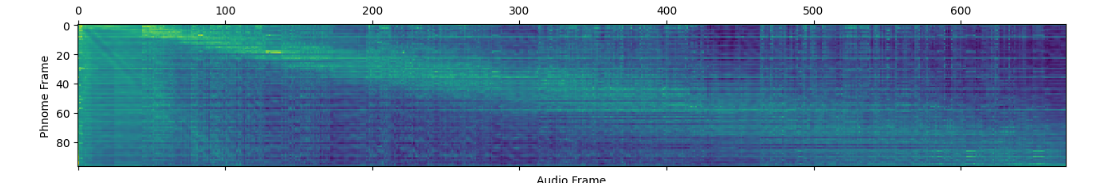}}
		\centerline{(a) Attention weight of VALL-E.}
	\end{minipage}
	\begin{minipage}[b]{1.0\linewidth}
		\centering
		\centerline{\includegraphics[width=14cm]{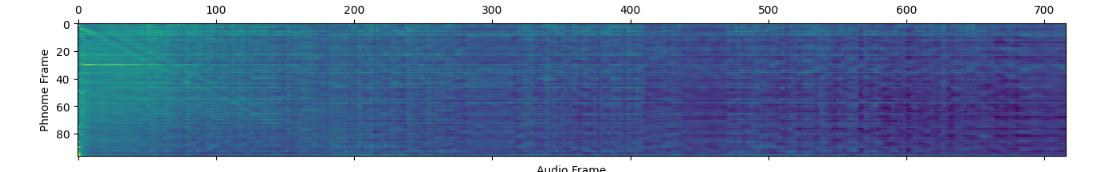}}
		\centerline{(b) Attention weight of VALL-E R with Monotonic Alignment.}
	\end{minipage}
	\begin{minipage}[b]{1.0\linewidth}
		\centering
		\centerline{\includegraphics[width=14cm]{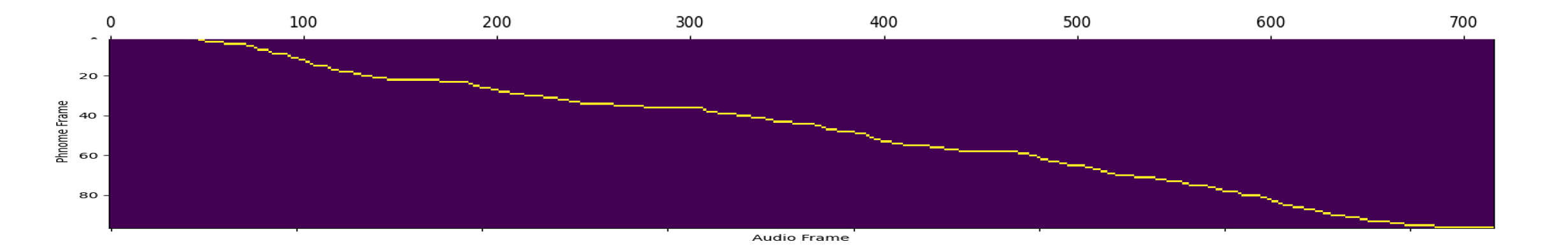}}
		\centerline{(c) Phoneme alignment path of VALL-E R.}
	\end{minipage}
	\caption{Visualization of Attention Weight between Phoneme Prompt and Generated Codec. The brighter the color, the greater the weight.}
	\label{fig:visual}
\end{figure*}


\paragraph{Visualization of Attention Weights}
To better demonstrate the mechanism behind the stability of our proposed VALL-E R model, we also conducted a deeper analysis through visualization. Given a speech sample $y$ and its transcription $x$, we calculated the average attention weights among 16 heads between the speech and text prompts within the first layer of the transformer during autoregressive inference process. And the weights are displayed in the Figure~\ref{fig:visual}. 

In the visualization for VALL-E, as depicted in Figure (a), we can observe a faint bright line along the diagonal. This indicates that during audio synthesis, VALL-E constantly needs to pay attention to which phoneme it is currently processing from the given prompt sequence. However, as the audio lengthens, the attention becomes increasingly scattered. This suggests that without explicit indications, the model becomes confused about which element to focus on, leading to misreadings, omissions, and repetitions. In contrast, for our proposed VALL-E R, depicted in Figure (b), there is no distinct bright line, and the attention is evenly distributed. This demonstrates that VALL-E R introduces explicit phoneme indications, allowing the model to focus on the global information of the text without worrying about the progress of reading.
Finally, in Figure (c), we provide the alignment path of phonemes in VALL-E R during the inference process. An uninterrupted and monotonous diagonal can be observed, which indicates that all phonemes are covered by the generated speech one by one, satisfying the characteristics of \textit{locality}, \textit{monotonicity}, and \textit{completeness} defined in Sec.~\ref{sec:inference}.


\begin{table}[h]
  \caption{Ablation study on Librispeech test-clean set. MA denotes monotonic alignment with phoneme. MC represents merged codec.}
  \label{tab:ablation_ma}
  \centering
  \begin{tabular}{lcccccc}
    \toprule
    Model     & MA &  MC & WER & Spk-Sim\\
    \midrule
    \multirow{2}{*}{VALL-E R}  & \checkmark  & \checkmark & 1.58 & 0.876 \\
             & \checkmark  & -  & 1.65 & 0.877 \\ \midrule
    VALL-E         & -  & - &  2.37 & 0.875 \\
             
    \bottomrule
  \end{tabular}
\end{table}


\paragraph{Ablation Study}
We also conduct ablation experiments to study the effectiveness of the methods we proposed, with the results presented in Table~\ref{tab:ablation_ma}. Initially, within VALL-E R, we replaced the merged codec (first layer downsample 2x) with the tradition codec. The results indicate that the merged codec does not adversely affect WER or Spk-sim while improving inference speed, thereby affirming the superiority of our proposed acceleration strategy. Subsequently, upon removing the monotonic phoneme strategy, the model regresses completely to VALL-E, leading to a significant deterioration in WER performance. This highlights the effectiveness of our proposed method in terms of robustness.

             

\section{Conclusion}
In this paper, we propose VALL-E R, a robust and efficient zero-shot text-to-speech model based on neural codec language models. We first employ codec-merging method to downsample the discrete codes to improve inference speed without affecting the sound quality. More importantly, we incorporate phoneme prediction into the training process and utilize monotonic alignment strategy during the inference. 
Experimental results demonstrate that VALL-E R can effectively model the correlation between phonemes and audio, significantly improve the robustness of speech synthesis, and endow the ability to flexibly control prosody. 

\section{Ethics Statement and Broader Impact}
Since VALL-E R  is capable of synthesize speech that preserves speaker identity, it may carry potential risks for misuse, such as spoofing voice identification or impersonating specific individuals. Our experiments were conducted under the premise that the user agree to be the target speaker in synthesised speech. If  the model be applied to unseen speakers in real-world scenarios, it will be imperative to establish a protocol that ensures the speaker's authorization for the use of their voice, as well as the integration of a model capable of detecting synthesized speech.
To mitigate such risks,a detection model could be developed to discriminate whether an audio clip has been synthesized by VALL-E R. Furthermore, We commit to adhering to the Microsoft AI Principles\footnote{https://www.microsoft.com/ai/responsible-ai} in practice when further developing the models.



\bibliography{neurips_2023}

\begin{thebibliography}{46}
\providecommand{\natexlab}[1]{#1}
\providecommand{\url}[1]{\texttt{#1}}
\expandafter\ifx\csname urlstyle\endcsname\relax
  \providecommand{\doi}[1]{doi: #1}\else
  \providecommand{\doi}{doi: \begingroup \urlstyle{rm}\Url}\fi

\bibitem[Achiam et~al.(2023)Achiam, Adler, Agarwal, Ahmad, Akkaya, Aleman, Almeida, Altenschmidt, Altman, Anadkat, et~al.]{achiam2023gpt}
Josh Achiam, Steven Adler, Sandhini Agarwal, Lama Ahmad, Ilge Akkaya, Florencia~Leoni Aleman, Diogo Almeida, Janko Altenschmidt, Sam Altman, Shyamal Anadkat, et~al.
\newblock Gpt-4 technical report.
\newblock \emph{arXiv preprint arXiv:2303.08774}, 2023.

\bibitem[Arik et~al.(2018)Arik, Chen, Peng, Ping, and Zhou]{arik2018neural}
Sercan Arik, Jitong Chen, Kainan Peng, Wei Ping, and Yanqi Zhou.
\newblock Neural voice cloning with a few samples.
\newblock \emph{Advances in neural information processing systems}, 31, 2018.

\bibitem[Battenberg et~al.(2020)Battenberg, Skerry-Ryan, Mariooryad, Stanton, Kao, Shannon, and Bagby]{battenberg2020location}
Eric Battenberg, RJ~Skerry-Ryan, Soroosh Mariooryad, Daisy Stanton, David Kao, Matt Shannon, and Tom Bagby.
\newblock Location-relative attention mechanisms for robust long-form speech synthesis.
\newblock In \emph{ICASSP 2020-2020 IEEE International Conference on Acoustics, Speech and Signal Processing (ICASSP)}, pages 6194--6198. IEEE, 2020.

\bibitem[Borsos et~al.(2023)Borsos, Marinier, Vincent, Kharitonov, Pietquin, Sharifi, Roblek, Teboul, Grangier, Tagliasacchi, et~al.]{borsos2023audiolm}
Zal{\'a}n Borsos, Rapha{\"e}l Marinier, Damien Vincent, Eugene Kharitonov, Olivier Pietquin, Matt Sharifi, Dominik Roblek, Olivier Teboul, David Grangier, Marco Tagliasacchi, et~al.
\newblock Audiolm: a language modeling approach to audio generation.
\newblock \emph{IEEE/ACM Transactions on Audio, Speech, and Language Processing}, 2023.

\bibitem[Brown et~al.(2020)Brown, Mann, Ryder, Subbiah, Kaplan, Dhariwal, Neelakantan, Shyam, Sastry, Askell, et~al.]{brown2020language}
Tom Brown, Benjamin Mann, Nick Ryder, Melanie Subbiah, Jared~D Kaplan, Prafulla Dhariwal, Arvind Neelakantan, Pranav Shyam, Girish Sastry, Amanda Askell, et~al.
\newblock Language models are few-shot learners.
\newblock \emph{Advances in neural information processing systems}, 33:\penalty0 1877--1901, 2020.

\bibitem[Chen et~al.(2020)Chen, Tan, Ren, Xu, Sun, Zhao, Qin, and Liu]{chen2020multispeech}
Mingjian Chen, Xu~Tan, Yi~Ren, Jin Xu, Hao Sun, Sheng Zhao, Tao Qin, and Tie-Yan Liu.
\newblock Multispeech: Multi-speaker text to speech with transformer.
\newblock \emph{arXiv preprint arXiv:2006.04664}, 2020.

\bibitem[Chen et~al.(2022{\natexlab{a}})Chen, Tan, Qi, Zhou, Li, and Wu]{chen2022v2c}
Qi~Chen, Mingkui Tan, Yuankai Qi, Jiaqiu Zhou, Yuanqing Li, and Qi~Wu.
\newblock V2c: visual voice cloning.
\newblock In \emph{Proceedings of the IEEE/CVF Conference on Computer Vision and Pattern Recognition}, pages 21242--21251, 2022{\natexlab{a}}.

\bibitem[Chen et~al.(2022{\natexlab{b}})Chen, Wang, Chen, Wu, Liu, Chen, Li, Kanda, Yoshioka, Xiao, et~al.]{chen2022wavlm}
Sanyuan Chen, Chengyi Wang, Zhengyang Chen, Yu~Wu, Shujie Liu, Zhuo Chen, Jinyu Li, Naoyuki Kanda, Takuya Yoshioka, Xiong Xiao, et~al.
\newblock Wavlm: Large-scale self-supervised pre-training for full stack speech processing.
\newblock \emph{IEEE Journal of Selected Topics in Signal Processing}, 16\penalty0 (6):\penalty0 1505--1518, 2022{\natexlab{b}}.

\bibitem[Chen et~al.(2018)Chen, Assael, Shillingford, Budden, Reed, Zen, Wang, Cobo, Trask, Laurie, et~al.]{chen2018sample}
Yutian Chen, Yannis Assael, Brendan Shillingford, David Budden, Scott Reed, Heiga Zen, Quan Wang, Luis~C Cobo, Andrew Trask, Ben Laurie, et~al.
\newblock Sample efficient adaptive text-to-speech.
\newblock \emph{arXiv preprint arXiv:1809.10460}, 2018.

\bibitem[Chen et~al.(2022{\natexlab{c}})Chen, Chen, Wu, Qian, Wang, Liu, Qian, and Zeng]{chen2022large}
Zhengyang Chen, Sanyuan Chen, Yu~Wu, Yao Qian, Chengyi Wang, Shujie Liu, Yanmin Qian, and Michael Zeng.
\newblock Large-scale self-supervised speech representation learning for automatic speaker verification.
\newblock In \emph{ICASSP 2022-2022 IEEE International Conference on Acoustics, Speech and Signal Processing (ICASSP)}, pages 6147--6151. IEEE, 2022{\natexlab{c}}.

\bibitem[Chiu and Raffel(2017)]{chiu2017monotonic}
Chung-Cheng Chiu and Colin Raffel.
\newblock Monotonic chunkwise attention.
\newblock \emph{arXiv preprint arXiv:1712.05382}, 2017.

\bibitem[Cooper et~al.(2020)Cooper, Lai, Yasuda, Fang, Wang, Chen, and Yamagishi]{cooper2020zero}
Erica Cooper, Cheng-I Lai, Yusuke Yasuda, Fuming Fang, Xin Wang, Nanxin Chen, and Junichi Yamagishi.
\newblock Zero-shot multi-speaker text-to-speech with state-of-the-art neural speaker embeddings.
\newblock In \emph{ICASSP 2020-2020 IEEE International Conference on Acoustics, Speech and Signal Processing (ICASSP)}, pages 6184--6188. IEEE, 2020.

\bibitem[Copet et~al.(2024)Copet, Kreuk, Gat, Remez, Kant, Synnaeve, Adi, and D{\'e}fossez]{copet2024simple}
Jade Copet, Felix Kreuk, Itai Gat, Tal Remez, David Kant, Gabriel Synnaeve, Yossi Adi, and Alexandre D{\'e}fossez.
\newblock Simple and controllable music generation.
\newblock \emph{Advances in Neural Information Processing Systems}, 36, 2024.

\bibitem[D{\'e}fossez et~al.(2022)D{\'e}fossez, Copet, Synnaeve, and Adi]{defossez2022high}
Alexandre D{\'e}fossez, Jade Copet, Gabriel Synnaeve, and Yossi Adi.
\newblock High fidelity neural audio compression.
\newblock \emph{arXiv preprint arXiv:2210.13438}, 2022.

\bibitem[Ding et~al.(2021)Ding, Yang, Hong, Zheng, Zhou, Yin, Lin, Zou, Shao, Yang, et~al.]{ding2021cogview}
Ming Ding, Zhuoyi Yang, Wenyi Hong, Wendi Zheng, Chang Zhou, Da~Yin, Junyang Lin, Xu~Zou, Zhou Shao, Hongxia Yang, et~al.
\newblock Cogview: Mastering text-to-image generation via transformers.
\newblock \emph{Advances in Neural Information Processing Systems}, 34:\penalty0 19822--19835, 2021.

\bibitem[Du et~al.(2024)Du, Guo, Wang, Yang, Niu, Wang, Zhang, Chen, and Yu]{du2024vall}
Chenpeng Du, Yiwei Guo, Hankun Wang, Yifan Yang, Zhikang Niu, Shuai Wang, Hui Zhang, Xie Chen, and Kai Yu.
\newblock Vall-t: Decoder-only generative transducer for robust and decoding-controllable text-to-speech.
\newblock \emph{arXiv preprint arXiv:2401.14321}, 2024.

\bibitem[Gong et~al.(2024)Gong, Wu, Li, Liu, Zhao, Chen, and Qian]{gong2024advanced}
Xun Gong, Yu~Wu, Jinyu Li, Shujie Liu, Rui Zhao, Xie Chen, and Yanmin Qian.
\newblock Advanced {{Long-Content Speech Recognition}} with {{Factorized Neural Transducer}}.
\newblock \emph{IEEE/ACM Transactions on Audio, Speech, and Language Processing}, 32:\penalty0 1--14, 2024.
\newblock ISSN 2329-9290, 2329-9304.
\newblock \doi{10.1109/TASLP.2024.3350893}.

\bibitem[Graves(2012)]{graves2012sequence}
Alex Graves.
\newblock Sequence transduction with recurrent neural networks.
\newblock \emph{arXiv preprint arXiv:1211.3711}, 2012.

\bibitem[Gulati et~al.(2020)Gulati, Qin, Chiu, Parmar, Zhang, Yu, Han, Wang, Zhang, Wu, et~al.]{gulati2020conformer}
Anmol Gulati, James Qin, Chung-Cheng Chiu, Niki Parmar, Yu~Zhang, Jiahui Yu, Wei Han, Shibo Wang, Zhengdong Zhang, Yonghui Wu, et~al.
\newblock Conformer: Convolution-augmented transformer for speech recognition.
\newblock \emph{arXiv preprint arXiv:2005.08100}, 2020.

\bibitem[Hao et~al.(2023)Hao, Zhou, Liu, Li, Hu, Wang, and Wei]{hao2023boosting}
Hongkun Hao, Long Zhou, Shujie Liu, Jinyu Li, Shujie Hu, Rui Wang, and Furu Wei.
\newblock Boosting large language model for speech synthesis: An empirical study.
\newblock \emph{arXiv preprint arXiv:2401.00246}, 2023.

\bibitem[He et~al.(2019)He, Deng, and He]{he2019robust}
Mutian He, Yan Deng, and Lei He.
\newblock Robust sequence-to-sequence acoustic modeling with stepwise monotonic attention for neural tts.
\newblock \emph{arXiv preprint arXiv:1906.00672}, 2019.

\bibitem[Jiang et~al.(2023)Jiang, Liu, Ren, He, Zhang, Ye, Wei, Wang, Yin, Ma, et~al.]{jiang2023mega}
Ziyue Jiang, Jinglin Liu, Yi~Ren, Jinzheng He, Chen Zhang, Zhenhui Ye, Pengfei Wei, Chunfeng Wang, Xiang Yin, Zejun Ma, et~al.
\newblock Mega-tts 2: Zero-shot text-to-speech with arbitrary length speech prompts.
\newblock \emph{arXiv preprint arXiv:2307.07218}, 2023.

\bibitem[Kharitonov et~al.(2023)Kharitonov, Vincent, Borsos, Marinier, Girgin, Pietquin, Sharifi, Tagliasacchi, and Zeghidour]{kharitonov2023speak}
Eugene Kharitonov, Damien Vincent, Zal{\'a}n Borsos, Rapha{\"e}l Marinier, Sertan Girgin, Olivier Pietquin, Matt Sharifi, Marco Tagliasacchi, and Neil Zeghidour.
\newblock Speak, read and prompt: High-fidelity text-to-speech with minimal supervision.
\newblock \emph{Transactions of the Association for Computational Linguistics}, 11:\penalty0 1703--1718, 2023.

\bibitem[Kim et~al.(2023)Kim, Lee, Chung, and Cho]{kim2023clam}
Jaehyeon Kim, Keon Lee, Seungjun Chung, and Jaewoong Cho.
\newblock Clam-tts: Improving neural codec language model for zero-shot text-to-speech.
\newblock In \emph{The Twelfth International Conference on Learning Representations}, 2023.

\bibitem[Kondratyuk et~al.(2023)Kondratyuk, Yu, Gu, Lezama, Huang, Hornung, Adam, Akbari, Alon, Birodkar, et~al.]{kondratyuk2023videopoet}
Dan Kondratyuk, Lijun Yu, Xiuye Gu, Jos{\'e} Lezama, Jonathan Huang, Rachel Hornung, Hartwig Adam, Hassan Akbari, Yair Alon, Vighnesh Birodkar, et~al.
\newblock Videopoet: A large language model for zero-shot video generation.
\newblock \emph{arXiv preprint arXiv:2312.14125}, 2023.

\bibitem[Le et~al.(2024)Le, Vyas, Shi, Karrer, Sari, Moritz, Williamson, Manohar, Adi, Mahadeokar, et~al.]{le2024voicebox}
Matthew Le, Apoorv Vyas, Bowen Shi, Brian Karrer, Leda Sari, Rashel Moritz, Mary Williamson, Vimal Manohar, Yossi Adi, Jay Mahadeokar, et~al.
\newblock Voicebox: Text-guided multilingual universal speech generation at scale.
\newblock \emph{Advances in neural information processing systems}, 36, 2024.

\bibitem[McAuliffe et~al.(2017)McAuliffe, Socolof, Mihuc, Wagner, and Sonderegger]{mcauliffe2017montreal}
Michael McAuliffe, Michaela Socolof, Sarah Mihuc, Michael Wagner, and Morgan Sonderegger.
\newblock Montreal forced aligner: Trainable text-speech alignment using kaldi.
\newblock In \emph{Interspeech}, volume 2017, pages 498--502, 2017.

\bibitem[Moss et~al.(2020)Moss, Aggarwal, Prateek, Gonz{\'a}lez, and Barra-Chicote]{moss2020boffin}
Henry~B Moss, Vatsal Aggarwal, Nishant Prateek, Javier Gonz{\'a}lez, and Roberto Barra-Chicote.
\newblock Boffin tts: Few-shot speaker adaptation by bayesian optimization.
\newblock In \emph{ICASSP 2020-2020 IEEE International Conference on Acoustics, Speech and Signal Processing (ICASSP)}, pages 7639--7643. IEEE, 2020.

\bibitem[Panayotov et~al.(2015)Panayotov, Chen, Povey, and Khudanpur]{panayotov2015librispeech}
Vassil Panayotov, Guoguo Chen, Daniel Povey, and Sanjeev Khudanpur.
\newblock Librispeech: an asr corpus based on public domain audio books.
\newblock In \emph{2015 IEEE international conference on acoustics, speech and signal processing (ICASSP)}, pages 5206--5210. IEEE, 2015.

\bibitem[Radford et~al.(2021)Radford, Kim, Hallacy, Ramesh, Goh, Agarwal, Sastry, Askell, Mishkin, Clark, et~al.]{radford2021learning}
Alec Radford, Jong~Wook Kim, Chris Hallacy, Aditya Ramesh, Gabriel Goh, Sandhini Agarwal, Girish Sastry, Amanda Askell, Pamela Mishkin, Jack Clark, et~al.
\newblock Learning transferable visual models from natural language supervision.
\newblock In \emph{International conference on machine learning}, pages 8748--8763. PMLR, 2021.

\bibitem[Raffel et~al.(2017)Raffel, Luong, Liu, Weiss, and Eck]{raffel2017online}
Colin Raffel, Minh-Thang Luong, Peter~J Liu, Ron~J Weiss, and Douglas Eck.
\newblock Online and linear-time attention by enforcing monotonic alignments.
\newblock In \emph{International conference on machine learning}, pages 2837--2846. PMLR, 2017.

\bibitem[Ramesh et~al.(2021)Ramesh, Pavlov, Goh, Gray, Voss, Radford, Chen, and Sutskever]{ramesh2021zero}
Aditya Ramesh, Mikhail Pavlov, Gabriel Goh, Scott Gray, Chelsea Voss, Alec Radford, Mark Chen, and Ilya Sutskever.
\newblock Zero-shot text-to-image generation.
\newblock In \emph{International conference on machine learning}, pages 8821--8831. Pmlr, 2021.

\bibitem[Siuzdak(2023)]{siuzdak2023vocos}
Hubert Siuzdak.
\newblock Vocos: Closing the gap between time-domain and fourier-based neural vocoders for high-quality audio synthesis.
\newblock \emph{arXiv preprint arXiv:2306.00814}, 2023.

\bibitem[Song et~al.(2024)Song, Chen, Wang, Ma, and Chen]{song2024ella}
Yakun Song, Zhuo Chen, Xiaofei Wang, Ziyang Ma, and Xie Chen.
\newblock Ella-v: Stable neural codec language modeling with alignment-guided sequence reordering.
\newblock \emph{arXiv preprint arXiv:2401.07333}, 2024.

\bibitem[Sotelo et~al.(2017)Sotelo, Mehri, Kumar, Santos, Kastner, Courville, and Bengio]{sotelo2017char2wav}
Jose Sotelo, Soroush Mehri, Kundan Kumar, Joao~Felipe Santos, Kyle Kastner, Aaron Courville, and Yoshua Bengio.
\newblock Char2wav: End-to-end speech synthesis.
\newblock 2017.

\bibitem[Tachibana et~al.(2018)Tachibana, Uenoyama, and Aihara]{tachibana2018efficiently}
Hideyuki Tachibana, Katsuya Uenoyama, and Shunsuke Aihara.
\newblock Efficiently trainable text-to-speech system based on deep convolutional networks with guided attention.
\newblock In \emph{2018 IEEE international conference on acoustics, speech and signal processing (ICASSP)}, pages 4784--4788. IEEE, 2018.

\bibitem[Tan et~al.(2021)Tan, Qin, Soong, and Liu]{tan2021survey}
Xu~Tan, Tao Qin, Frank Soong, and Tie-Yan Liu.
\newblock A survey on neural speech synthesis.
\newblock \emph{arXiv preprint arXiv:2106.15561}, 2021.

\bibitem[Touvron et~al.(2023)Touvron, Martin, Stone, Albert, Almahairi, Babaei, Bashlykov, Batra, Bhargava, Bhosale, et~al.]{touvron2023llama}
Hugo Touvron, Louis Martin, Kevin Stone, Peter Albert, Amjad Almahairi, Yasmine Babaei, Nikolay Bashlykov, Soumya Batra, Prajjwal Bhargava, Shruti Bhosale, et~al.
\newblock Llama 2: Open foundation and fine-tuned chat models.
\newblock \emph{arXiv preprint arXiv:2307.09288}, 2023.

\bibitem[Vasquez and Lewis(2019)]{vasquez2019melnet}
Sean Vasquez and Mike Lewis.
\newblock Melnet: A generative model for audio in the frequency domain.
\newblock \emph{arXiv preprint arXiv:1906.01083}, 2019.

\bibitem[Wang et~al.(2023{\natexlab{a}})Wang, Chen, Wu, Zhang, Zhou, Liu, Chen, Liu, Wang, Li, et~al.]{wang2023neural}
Chengyi Wang, Sanyuan Chen, Yu~Wu, Ziqiang Zhang, Long Zhou, Shujie Liu, Zhuo Chen, Yanqing Liu, Huaming Wang, Jinyu Li, et~al.
\newblock Neural codec language models are zero-shot text to speech synthesizers.
\newblock \emph{arXiv preprint arXiv:2301.02111}, 2023{\natexlab{a}}.

\bibitem[Wang et~al.(2023{\natexlab{b}})Wang, Zhou, Zhang, Wu, Liu, Gaur, Chen, Li, and Wei]{wang2023viola}
Tianrui Wang, Long Zhou, Ziqiang Zhang, Yu~Wu, Shujie Liu, Yashesh Gaur, Zhuo Chen, Jinyu Li, and Furu Wei.
\newblock Viola: Unified codec language models for speech recognition, synthesis, and translation.
\newblock \emph{arXiv preprint arXiv:2305.16107}, 2023{\natexlab{b}}.

\bibitem[Xin et~al.(2024)Xin, Tan, Shen, Ju, Yang, Wang, Takamichi, Saruwatari, Liu, Li, et~al.]{xin2024rall}
Detai Xin, Xu~Tan, Kai Shen, Zeqian Ju, Dongchao Yang, Yuancheng Wang, Shinnosuke Takamichi, Hiroshi Saruwatari, Shujie Liu, Jinyu Li, et~al.
\newblock Rall-e: Robust codec language modeling with chain-of-thought prompting for text-to-speech synthesis.
\newblock \emph{arXiv preprint arXiv:2404.03204}, 2024.

\bibitem[Yu et~al.(2024)Yu, Simig, Flaherty, Aghajanyan, Zettlemoyer, and Lewis]{yu2024megabyte}
Lili Yu, D{\'a}niel Simig, Colin Flaherty, Armen Aghajanyan, Luke Zettlemoyer, and Mike Lewis.
\newblock Megabyte: Predicting million-byte sequences with multiscale transformers.
\newblock \emph{Advances in Neural Information Processing Systems}, 36, 2024.

\bibitem[Zeghidour et~al.(2021)Zeghidour, Luebs, Omran, Skoglund, and Tagliasacchi]{zeghidour2021soundstream}
Neil Zeghidour, Alejandro Luebs, Ahmed Omran, Jan Skoglund, and Marco Tagliasacchi.
\newblock Soundstream: An end-to-end neural audio codec.
\newblock \emph{IEEE/ACM Transactions on Audio, Speech, and Language Processing}, 30:\penalty0 495--507, 2021.

\bibitem[Zhang et~al.(2023)Zhang, Zhou, Wang, Chen, Wu, Liu, Chen, Liu, Wang, Li, et~al.]{zhang2023speak}
Ziqiang Zhang, Long Zhou, Chengyi Wang, Sanyuan Chen, Yu~Wu, Shujie Liu, Zhuo Chen, Yanqing Liu, Huaming Wang, Jinyu Li, et~al.
\newblock Speak foreign languages with your own voice: Cross-lingual neural codec language modeling.
\newblock \emph{arXiv preprint arXiv:2303.03926}, 2023.

\bibitem[Zhao et~al.(2023)Zhao, Kong, Liang, Zhu, Kuang, and Wu]{zhao2023clap}
Tianqi Zhao, Ming Kong, Tian Liang, Qiang Zhu, Kun Kuang, and Fei Wu.
\newblock Clap: Contrastive language-audio pre-training model for multi-modal sentiment analysis.
\newblock In \emph{Proceedings of the 2023 ACM International Conference on Multimedia Retrieval}, pages 622--626, 2023.

\end{thebibliography}
\bibliographystyle{plainnat}

\clearpage

\end{document}